\documentclass[10pt,twocolumn,letterpaper]{article}

\usepackage{cvpr}              
\usepackage[accsupp]{axessibility}

\usepackage{siunitx}
\usepackage{multirow}

\definecolor{cvprblue}{rgb}{0.21,0.49,0.74}
\usepackage[pagebackref,breaklinks,colorlinks,allcolors=cvprblue]{hyperref}

\usepackage{xcolor}

\newcommand{\citehere}[1]{\textcolor{red}{(CITE)}}

\def\paperID{60} 
\def\confName{EarthVision}
\def\confYear{2026}

\title{A Proxy Consistency Loss for Grounded Fusion of Earth Observation and Location Encoders
}

\author{Zhongying Wang, Kevin Lane, Levi Cai, Morteza Karimzadeh, Esther Rolf\\
University of Colorado, Boulder\\
{\tt\small \{zhongying.wang,kevin.lane,levi.cai,karimzadeh,esther.rolf\}@colorado.edu}}

\begin{document}
\maketitle
\begin{abstract}
Supervised learning with Earth observation inputs is often limited by the sparsity of high-quality labeled or in-situ measured data to use as training labels. With the abundance of geographic data products, in many cases there are variables correlated with -- but different from -- the variable of interest that can be leveraged. 
%
We integrate such proxy variables within a geographic prior via a trainable location encoder and introduce a proxy consistency loss (PCL) formulation to imbue proxy data into the location encoder. 
The first key insight behind our approach is to use the location encoder as an agile and flexible way to learn from abundantly available proxy data which can be sampled independently of training label availability. 
Our second key insight is that we will need to regularize the location encoder appropriately to achieve performance and robustness with limited labeled data. 
Our experiments on air quality prediction and poverty mapping show that integrating proxy data implicitly through the location encoder outperforms using both as input to an observation encoder and fusion strategies that use frozen, pretrained location embeddings as a geographic prior. Superior performance for in-sample prediction shows that the PCL can incorporate rich information from the proxies, and superior out-of-sample prediction shows that the learned latent embeddings help generalize to areas without training labels. 
\end{abstract}

\section{Introduction}
One of the most pervasive challenges in building task-specific machine learning models with Earth observation (EO) data is the limited availability of high-quality labeled data. For example, in environmental monitoring, in-situ measurements of air or water quality are isolated to sparse points where sensors are placed \cite{SABERIOON2020106236,considine2023,Schneide2020pm}. In agriculture and development economics, data are often obtained from surveys conducted over a small set of clusters or sensor networks \cite{yeh2020using, Xu2022SMAP}. 
The sparse coverage of labeled data, which is costly to collect, makes it difficult to train machine learning models that generalize over large areas.

Here, we aim to tap into the potential of more abundant proxy data relevant to---but not exactly matching---the task at hand, to better condition predictive models. For example, \citet{jean2016combining} employed transfer learning to leverage widely available VIIRS nighttime luminosity data to pretrain their model before fine-tuning it for predicting poverty from limited cluster labels. Other approaches have shown that proxy data can be used for multi-task pretraining during self-supervised learning of EO foundation models (e.g., \cite{wang2024mtp,nedungadi2024mmearth,brown2025alphaearth,sosa2025multimae}), and even simply stacking proxy data as additional input bands can aid performance in label-scarce and out-of-domain settings \cite{rao2025using}. 

Newer approaches have used location encoder models that map location inputs into higher dimensional representations (embeddings) \cite{klemmer2025earth, vivanco2023geoclip} to condition task-specific observation-based models. Location encoders are a relatively agile and lightweight form of geographic conditioning.  Using frozen pretrained location encoders has been shown to benefit task-specific predictions in label-scarce settings and in settings of geographic domain shift, including species estimation \cite{Aodha_2019_ICCV}, air quality estimation \cite{karimzadeh2025performance}, and environmental and socio-economic variable mapping \cite{rao2025using,crasto2025robustness}. However, specializing high dimensional location encoders to task-specific models remains difficult (aside from the cost of pretraining location encoders), since unfreezing weights of the location encoder can lead to overfitting to training points \cite{rao2025using}. 
We hypothesize that there should be a way to spatially regularize trainable location encoders for better fusion in task-specific modeling.

\begin{figure*}[t]
    \centering
    \includegraphics[width=1\linewidth, trim=18em 33.8em 25em 1em,clip]{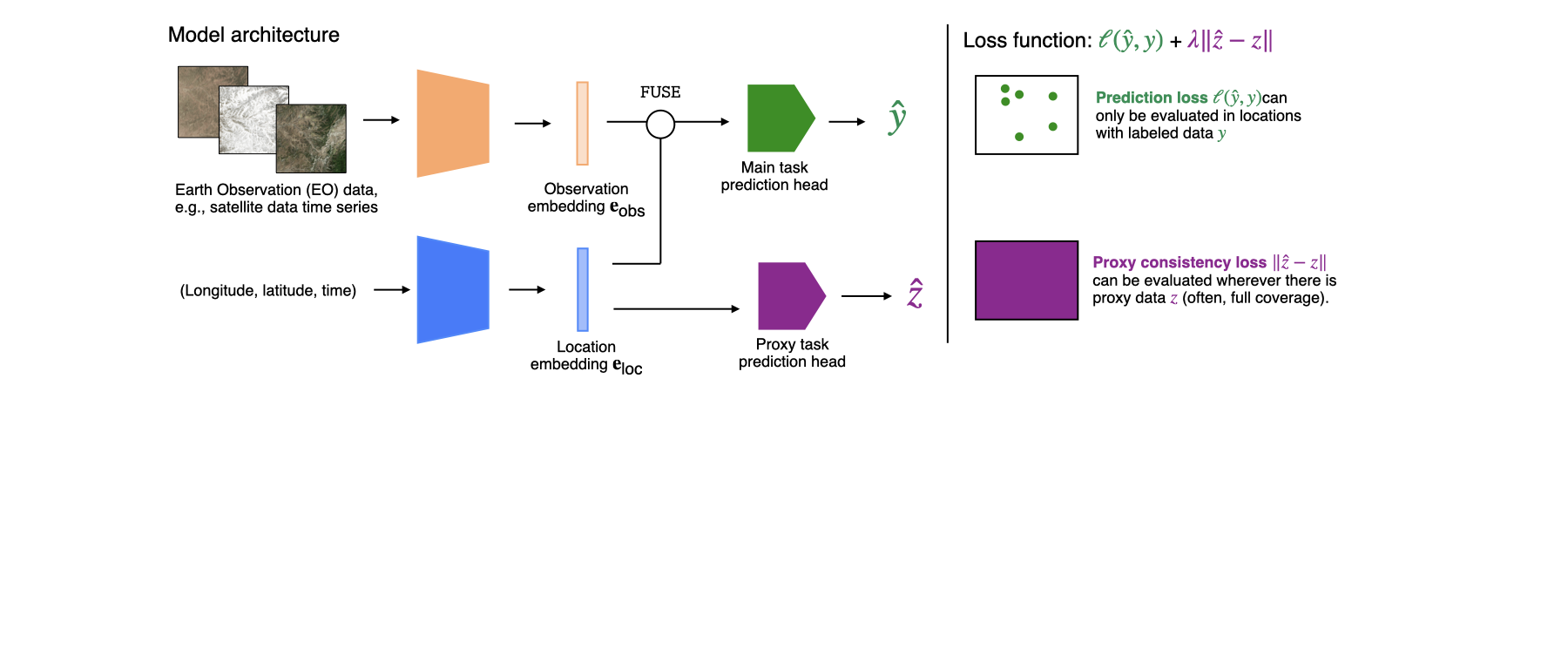}
    \caption{Our model architecture fuses a trainable location encoder (blue) with an EO-based encoder model for prediction of a supervised learning task $y$, while the location encoder must also be able to construct related proxy tasks $z$. Our proposed proxy consistency loss is incorporated in a multi-task loss, where the primary prediction task is a location-grounded prediction task (green), and an auxiliary proxy task (purple) is predicted using just the location encoder outputs. The overall objective function minimized during training is a weighted combination of the standard prediction loss and the proxy consistency loss which regularizes the location encoder.}
    \label{fig:explanatory_figure}
\end{figure*}

We propose a new \textit{proxy consistency loss} (PCL) to train geographically-conditioned predictors built on EO data. This takes the form of a multi-task framework where a location-time encoder is jointly optimized with (i) a point-observation prediction loss, and (ii) a proxy grounding loss that constructs the proxy continuous surface(s) from the location encoder, enabling ``everywhere" supervision without requiring more labels that are otherwise costly or impossible to collect (\Cref{fig:explanatory_figure}).  The intuition for our approach is that the same latent space that supports generalizable prediction of a specific task should also capture information about related proxy variables that are available more abundantly.
Thus, the auxiliary proxy loss acts as a form of geospatial regularizer, and a form of contextualization to ground the location embeddings in a task-specific, relevant high dimensional space. 


We demonstrate that proxy-consistent location encoders outperform both naïve proxy stacking and frozen pretrained location encoder baselines, especially under spatial domain shift. Additionally, we show that proxy-consistent training allows the model to specialize to the prediction task without overfitting to sparse training points, providing a more robust geographic prior for EO prediction problems.
While our experiments here focus on PM$_{2.5}$ and a poverty prediction benchmark, the proposed framework is applicable to other EO tasks—such as soil moisture estimation or water quality monitoring—where sparse point observations can be complemented by spatially continuous proxy variables.

\section{Related Work}
\label{sec:related work}

Our work ties together two key ideas in computer vision for Earth observation: using abundant proxy data to better condition the latent representation of an observation-based prediction model, and fusing location inputs for geographic conditioning of Earth observation-based predictions.

\subsection{Using proxy and auxiliary data}
Several prior works demonstrate the benefit of using proxy data---related to, but distinct from---the key prediction task at hand. Different approaches have rested on one core intuition: using the proxy data to learn a better latent representation of the Earth observation encoder, to be used for supervised learning with sparse labels. For the task of poverty mapping, \citet{jean2016combining} used transfer learning to first train a model on abundant nighttime luminosity data before fine-tuning the models for poverty prediction where only sparsely clustered labels are available. \citet{rao2025using} showed that even a simple approach that stacks proxy data like landcover maps as additional input to an Earth observation encoder can improve multiple tasks in machine learning with Earth observation data, with significant gains in label-scarce settings and out-of-domain prediction. \citet{Jalayer2025AutoICE} showed that stacking very coarse resolution passive microwave observations can improve the mapping of sea ice with high-resolution synthetic aperture radar.  


\subsection{Conditioning EO models with location encoders}
\label{sec:related_work_conditional_models}


Location can provide valuable context for machine learning models to learn patterns in spatiotemporal data.  Key to these geographically contextualized models is geospatial prior belief, a function of latitude and longitude position, that when fused with an observation (such as image) encoder, could improve prediction, such as better contextualized classifications of species from images \cite{Aodha_2019_ICCV,chu2019geo}. More recently, pretrained location encoders such as GeoCLIP \cite{vivanco2023geoclip}, SatCLIP \cite{klemmer2025satclip}, and CSP \cite{mai2023csp}, have been used to condition geospatial prediction models with latent embeddings learned via self-supervised learning. Using these pretrained, general purpose embeddings as geographic priors has been shown to aid performance in supervised learning of air quality prediction \cite{karimzadeh2025performance}, species classification, poverty mapping, and landcover classification \cite{wang2025high,crasto2025robustness}. Beyond supervised learning tasks, Earth embeddings have been shown to be a good latent space by which to condition retrieval \cite{dhakal2025range} and synthetic satellite image generation models \cite{sastry2024geosynth}.

We devise a novel method for training task-specific geographic priors by shaping the latent space of location embeddings to enforce consistency with relevant proxy variables. In doing so, we hope to overcome limitations of past approaches: general purpose pretrained embeddings may be too general to be related to the task at hand, and training task-specific location encoders without other supervision can lead to overfitting to sparse training locations \cite{rao2025using}.
Recent work suggests this may be possible; \citet{crasto2025robustness} shows that an auxiliary categorical domain prediction loss can improve spatial domain-robustness of geographically conditioned models. We expand significantly on this idea by using continuous fields as proxy targets which can be sampled across the entire study domain during training to generate task-specific, but geographically generalizable, location embeddings for observation-location fusion.

\section{Methods}
\label{sec:methods}
\subsection{Location encoder fusion with proxy tasks}

Following past work (see related work in \Cref{sec:related_work_conditional_models}), our model architecture fuses the outputs of an observation encoder and a location encoder at the embedding level. The observation encoder takes in observed data, like time series of satellite images, and outputs an embedding $\mathbf{e}_{\textrm{obs}} \in \mathbb{R}^{d_1}$. The location encoder takes in locations, e.g., (longitude, latitude, time), and outputs another embedding $\mathbf{e}_{\textrm{loc}} \in \mathbb{R}^{d_2}$. In this work, because we are focused on regression tasks, we fuse the two models by concatenating the output embeddings of each encoder $\mathtt{fuse}(\mathbf{e}_{\textrm{obs}},\mathbf{e}_{\textrm{loc}}) = \mathtt{concat}(\mathbf{e}_{\textrm{obs}}, \mathbf{e}_{\textrm{loc}}) = \begin{bmatrix} \mathbf{e}_{\textrm{obs}}^\top & \mathbf{e}_{\textrm{loc}}^\top \end{bmatrix}^\top$. The final prediction $\hat{y} = f(\mathtt{fuse}(\mathbf{e}_{\textrm{obs}},\mathbf{e}_{\textrm{loc}}))$ is a small prediction head that takes as input the concatenated embeddings (we use a small multi-layer perceptron, MLP). 

The key insight of our work is to train the location encoder with a multi-task prediction by using proxy variables $z$
related to the main prediction task $y$ (purple branch in \Cref{fig:explanatory_figure}). A proxy prediction task is one that is related to the task at hand, often correlated but not perfectly predictive of the main task. For example, nighttime luminosity has been shown to be a proxy of relative wealth in parts of the world \cite{jean2016combining}, and low-resolution air quality estimates are available across the US to supplement high-fidelity in-situ measurements. For our purposes, good proxy tasks are those that contain signal to the main prediction task, but are available with full coverage across the domain of interest.

We hypothesize that the multi-task objective will prevent unfrozen location encoders from overfitting  to the specific locations in the training set for the main task $y$, which has been shown to be a problem in past work \cite{rao2025using}. Because only the location encoder contributes to the proxy prediction, we can sample proxy points from across our region of interest during training, rather than being constrained to the locations where labeled data is available.
Thus, the additional proxy data should help condition the location encoder to produce more globally informative geographic priors, which can be critical in label-scarce settings and spatially out-of-domain prediction.

\subsection{Proxy consistency loss}

The prediction loss is a standard supervised loss $\mathcal{L}^{\textrm{pred}} = \ell(\hat{y},y)$. 
 We use mean squared error (MSE) throughout since our tasks are regression tasks, but one could easily use another loss, e.g., cross entropy for binary outputs.

The proxy consistency loss (PCL) we propose measures how well the location encoder can construct the proxy variables via $\hat{z} = g(\mathbf{e}_{\textrm{loc}})$, where $g$ is a prediction head, e.g., a small MLP.  For scalar-valued proxies:
\begin{align*}
    \mathcal{L}^{\textrm{proxy-consistency}} = \|\hat{z}-z\|_2^2~.
\end{align*}
For vector-valued proxies $\mathbf{z} \in \mathbb{R}^m$ with multiple components, we can define a generalized version $\mathcal{L}^{\textrm{proxy-consistency}} = (\mathbf{\hat{z}}-\mathbf{z})^\top \Lambda (\mathbf{\hat{z}}-\mathbf{z})$, where $\Lambda = \textrm{diag}(\lambda_1, \ldots \lambda_m)$ denotes the different weights for the $m$ different proxy variables.

The overall loss we propose is simply a weighted sum of the prediction task loss and the PCL:
\begin{align}
\mathcal{L} = \mathcal{L}^{\textrm{pred}} + \lambda \mathcal{L}^{\textrm{proxy-consistency}}~.
    \label{eq:overall_loss}
\end{align}

The prediction component of the loss $\mathcal{L}^{\textrm{pred}} = \ell(\hat{y},y)$ back-propagates through all trainable parameters in both the observation encoder and the location encoder. The proxy consistency loss, $\mathcal{L}^{\textrm{proxy-consistency}}$, on the other hand, only directly influences the trainable parameters of the location encoder. Thus, it acts as a sort of regularizer in that the latent embedding space generated from the location encoder must be useful both for the prediction task and the construction of related proxy data. Of course, this will \emph{indirectly} influence the optimal parameters for the observation encoder $\mathcal{L}^{\textrm{pred}}$, hopefully allowing it to specialize better to the task at hand, with a more reliable geographic prior from the location encoder.

\subsection{Sampling the proxy data separately}\label{sec:methods_pcl_sampling}
Since proxy variables will often be available across the entire region of interest, they can be used to condition the location embeddings everywhere, not just where the location encoder can be trained. To realize this, we sample batches of proxy variables separately from labeled observations during training. This allows us to set our own sampling scheme to train the location encoder; specifically, we can sample uniformly at random (UAR) over the geography of interest, which is likely desirable for training location encoders \cite{klemmer2025satclip}. 
We denote by $\rho$ the proxy sampling ratio: for each minibatch of $B$ labeled samples, we draw an additional minibatch of $\rho B$ proxy samples (UAR over space--time), so $\rho = \frac{|\mathcal{B}_{\text{proxy}}|}{|\mathcal{B}_{\text{sup}}|}$.

\section{Experimental Setup}
\subsection{Datasets and Tasks}

We focus primarily on the PM$_{2.5}$ prediction task for our experiments. To understand the generalizability of our method, we also study a wealth-index prediction task using the established poverty mapping benchmark in SustainBench \cite{yeh2021sustainbench}. More details on the datasets and training process are given in the supplementary material (SM).

\paragraph{Air quality prediction} For the prediction task, we use a contiguous U.S. (CONUS)-scale daily PM$_{2.5}$ mapping benchmark adapted from \citet{wang2025high}, 
where \textit{ground observations} from the U.S. EPA Air Quality System (AQS) are the primary learning target (\Cref{fig:epa-checkerboard-split} in the SM). We use measurements from 2017-2018, comprising approximately 
400K samples across 996 unique measurement sites. Following \cite{wang2025high}, the observation encoder inputs contain 18 satellite-derived \textit{columnar} aerosol measurements and environmental features. Deviating slightly from \cite{wang2025high}, we exclude the inverse-distance-weighted EPA PM$_{2.5}$ feature to isolate the effect of the location embeddings on spatial smoothing. 
For the proxy variable, we use the reanalysis of total PM$_{2.5}$ from the National Center for Atmospheric Research (NCAR) CONUS air quality reanalysis, a 12 km, hourly dataset spanning 2005-2018 \cite{kumar2025long}. While the prediction and proxy targets in this case are both related to PM$_{2.5}$, the prediction targets are strictly real ground-truth observations (measuring \textit{surface-level} PM$_{2.5}$, whereas the proxy target is a data assimilated product that provides globally distributed estimates using physical models based on coarse resolution observation of columns of aerosols. 
In separate ablations (see \Cref{subsec:pcl_ablations}), we also explore using other reanalysis variables as multi-channel proxy targets, including those that are not measurements of PM$_{2.5}$ but have a physical relationship to PM$_{2.5}$ formation. 

\paragraph{Poverty mapping} We use the poverty mapping benchmark established by SustainBench \cite{yeh2021sustainbench}. 
The prediction target is an asset index, as measured by the Demographic and Health Surveys (DHS), and the dataset uses clustered survey responses spanning from 1997 to 2018 that covers 52 countries, mainly focused in Africa. We use the provided 3-year composite of Landsat imagery \cite{wulder2022fifty} as the Earth observation encoder inputs. To best utilize this dataset for our study, we focus only on samples from Africa.  For our proxy target, we use patch-level heuristics from annual composites of VIIRS night lights images \cite{elvidge2017viirs}.  To ensure temporal consistency across our supervised task and our proxy target, we only train and test with samples ranging from 2013 to 2018.  To ensure consistency with our PM$_{2.5}$ task, we also do not use the ground-level images made available in the SustainBench task.

The air quality and poverty mapping tasks have key differences. They represent different spatial scales: the air quality task covers the CONUS across 996 EPA stations, and the poverty mapping task spans 19391 samples from 30 countries in Africa. The types of proxy variables also vary across tasks. The proxy target for air quality monitoring includes a coarser reanalysis of the primary target variable (PM$_{2.5}$) as well as ablations for other related (but not the same) physical variables, while the proxy variable for the poverty mapping task is a remotely-sensed product which is correlated but fundamentally different from the primary targets. 

\subsection{Spatial train/test splits}

For the air quality prediction task, we generate spatial train/validation and test splits to study in-sample and out-of-sample performance of our method.
To evaluate geographic in-sample performance, we randomly assign 50\% of EPA monitoring sites to train and the remaining 50\% to testing. Because test locations can be geographically close to training locations, the resulting evaluation primarily reflects interpolation within the sampling space. To evaluate geographic out-of-sample performance, we follow the systematic checkerboard protocol introduced in \cite{rolf2021generalizable}: we partition the study region into a checkerboard grid with square side length $\delta$ (in degrees), train on samples falling in alternating squares, and evaluate on the held-out squares (\Cref{fig:epa-checkerboard-split} in SM). We reported results including variation due to shifting and reversing the checkerboard pattern.





For the poverty mapping task, we start with the train/test split defined by TorchSpatial \cite{wu2024torchspatial}, where each country's samples are split into training and testing sets, then filter for samples recorded in African countries from 2013 to 2018.
This yields 15519 training samples and 3872 testing samples. To create the proxy variable, we randomly sample VIIRS \cite{elvidge2017viirs} annual night lights composite images ranging from 2013 to 2018 covering the continent of Africa.

\subsection{Model architectures and training procedure}

We fuse a location embedding with the observation embedding (by concatenation) prior to the final regression head, as shown in \Cref{fig:explanatory_figure}. 
For the trainable location encoders, we utilize the GeoCLIP-style \cite{vivanco2023geoclip} location encoding built on Equal Earth projection and multi-scale random Fourier features (RFF), with weights randomly initialized. To include the temporal aspect of position, we use an RFF encoder for day-of-year for the air quality prediction task, where seasonal and cyclical trends are critical, and a simple MLP for year-encoding for the poverty mapping task.

To incorporate the PCL regularization, we add a small decoder head on top of the location-time embedding to reconstruct the proxy target; the supervised loss updates the primary task-head, the observation and location encoders, and fusion layer, while the proxy-consistency loss updates the location-time encoder (and proxy task-head). 
We also train a two-stage variant for comparison, where we first pretrain the location-time encoder on proxy targets, then freeze it and train the observation-fusion model as above. 

\paragraph{Air quality prediction} Our observation-based backbone follows \cite{wang2025high}: a three-layer bidirectional Long Short-Term Memory (LSTM) \cite{graves2012long, schuster1997bidirectional} with Luong attention \cite{luong2015effective} produces a fixed-length sequence representation, which is mapped to daily PM$_{2.5}$ via a shallow MLP regression head. We adopted this architecture based on evaluations in \cite{wang2025high}, which showed state-of-the-art performance for this task.  

\paragraph{Poverty mapping}
Our observation-based encoder is a simple ResNet18 \cite{he2016deep} that has been initialized with pre-trained weights from SSL4E0-L \cite{stewart2023ssl4eo} by using the MoCo \cite{he2020momentum} self-supervised objective on Landsat imagery.  The proxy target is a tensor of patch-level details of VIIRS night light values, containing $(\text{centroid}, \text{mean}, \text{min}, \text{max})$.

\subsection{Baselines}
We compare our proposed method to several baselines. First, we compute a proxy-only linear regression to show that the proxy variable is not fully predictive of the primary target. Second, we train an observation encoder without location inputs, using the same observation encoder backbone as we use in the fused models. Third, we include proxy data as a raster stacked as an additional input layer to the observation encoder (with no location encoder fusion), by adding an additional input band to the observation encoder. The first three baselines provide baselines for different methods for training Earth observation models with and without proxy data. We also test how our task-specific location embeddings compare with \textit{Frozen pretrained LE fusion}, using general purpose pretrained location encoders (e.g., GeoCLIP \cite{vivanco2023geoclip} and Climplicit \cite{dollinger2025climplicit}).

\subsection{Evaluation metrics}
We evaluate performance with standard regression metrics: mean absolute error (MAE) and root mean squared error (RMSE), and $R^2$ score (fraction of variation in the labels explained by variation in the predictions). We additionally report mean bias error (MBE) $\frac{1}{N} \sum_{i=1}^{N}(\hat{y}_{i}-y_{i})$ for air quality prediction to quantify systematic under-/over-estimation.

\section{Results}

\subsection{Spatial Interpolation Performance}

\begin{table*}[htb]
\centering
\small
\setlength{\tabcolsep}{3pt}
\begin{tabular}{lccc ccc}
\toprule
\textbf{Method} &
\multicolumn{3}{c}{\textbf{UAR 50/50}} &
\multicolumn{3}{c}{\textbf{Checkerboard} ($\boldsymbol{\delta}=8^{\circ}$)} \\
\cmidrule(lr){2-4}\cmidrule(lr){5-7}
&
$\boldsymbol{R^{2}}$ &
\textbf{RMSE} (\si[per-mode=symbol]{\micro\gram\per\cubic\meter}) &
\textbf{MBE} (\si[per-mode=symbol]{\micro\gram\per\cubic\meter}) &
$\boldsymbol{R^{2}}$ &
\textbf{RMSE} (\si[per-mode=symbol]{\micro\gram\per\cubic\meter}) &
\textbf{MBE} (\si[per-mode=symbol]{\micro\gram\per\cubic\meter}) \\
\midrule

\multicolumn{7}{l}{\textit{Baselines}} \\
Proxy-only regression &
0.230 & 6.101 & 0.049 &
$0.215 \pm 0.010$ & $6.086 \pm 0.283$ & $0.001 \pm 0.258$ \\
Obs. encoder only &
$0.598 \pm 0.001$ & $4.407 \pm 0.005$ & $-0.045 \pm 0.021$ &
$0.331 \pm 0.018$ & $5.603 \pm 0.246$ & $-0.405 \pm 0.125$ \\
Proxy-stacked obs. encoder &
$0.610 \pm 0.002$ & $4.344 \pm 0.013$ & $-0.121 \pm 0.025$ &
$0.355 \pm 0.021$ & $5.496 \pm 0.221$ & $-0.294 \pm 0.132$ \\

\midrule
\multicolumn{7}{l}{\textit{Frozen location-encoder fusion}} \\
GeoCLIP &
$0.631 \pm 0.002$ & $4.222 \pm 0.012$ & $-0.178 \pm 0.016$ &
\textit{0.383} $\pm$ 0.018 & $5.393 \pm 0.235$ & $-0.573 \pm 0.095$ \\
Climplicit &
$0.642 \pm 0.002$ & $4.161 \pm 0.009$ & $-0.026 \pm 0.014$ &
$0.347 \pm 0.022$ & $5.536 \pm 0.278$ & $-0.502 \pm 0.160$ \\
Proxy pretraining (ours) &
\textit{0.666} $\pm$ 0.002 & \textit{4.016} $\pm$ 0.013 & $-0.124 \pm 0.014$ &
\textit{0.383} $\pm$ 0.017 & \textit{5.392} $\pm$ 0.259 & $-0.320 \pm 0.138$ \\

\midrule
\multicolumn{7}{l}{\textit{Trained location-encoder fusion}} \\
without PCL &
$0.655 \pm 0.001$ & $4.082 \pm 0.007$ & $-0.006 \pm 0.014$ &
$0.352 \pm 0.028$ & $5.522 \pm 0.287$ & $-0.663 \pm 0.111$ \\
PCL ($\lambda=0.2$, $\rho = 16$) &
$\mathbf{0.671} \pm 0.001$ & $\mathbf{3.988} \pm 0.006$ & $-0.024 \pm 0.012$ &
$\mathbf{0.390} \pm 0.024$ & $\mathbf{5.356} \pm 0.273$ & $-0.227 \pm 0.131$ \\
\bottomrule
\end{tabular}
\caption{\textbf{Performance summary for the air quality task on the UAR 50/50 spatial split and the systematically offset checkerboard split ($\delta=8^{\circ}$).}
For UAR 50/50, metrics are reported as mean over 5 random seeds $\pm$ standard error (SE); proxy-only regression is deterministic (no SE).
For checkerboard, we evaluate each method over 8 partitions (4 spatial offsets $\times$ train/test swap) and report mean $\pm$ SE across partitions.
Best results are in bold, second best are italicized.}
\label{tab:pm2.5_uar_checkerboard_r2_rmse_mbe}
\end{table*}

\Cref{tab:pm2.5_uar_checkerboard_r2_rmse_mbe} and \Cref{tab:app_pm25_mae_time} show in-sample performance on the UAR 50/50 spatial split. The \textit{proxy-only} linear regressor using the NCAR reanalysis performs poorly, indicating substantial residual between the 12 km chemical transport model outputs and point-scale, real AQS targets. Naively stacking the reanalysis proxy as an additional input provides only marginal gains over the observation encoder only. 

In contrast, fusing frozen, pretrained location encoders yields consistent improvements. This aligns with prior evidence that location encoders provide useful geographic context for dynamic PM$_{2.5}$ estimation \cite{karimzadeh2025performance}. Using frozen Climplicit embeddings outperforms using frozen GeoCLIP embeddings, likely due to its spatiotemporal pretraining. Training a task-specific location-time encoder from scratch is competitive, and adding our proposed proxy consistency loss achieves the best overall performance, with a $12.2\%$ relative improvement in $R^{2}$ and a $9.5\%$ reduction in RMSE over the observation-only baseline. Notably, joint PCL training outperforms the two-stage proxy-pretraining while avoiding an additional training phase. The performance is consistent across $R^2$, RMSE, and MAE. While the no-PCL fusion exhibits the lowest overall bias, our PCL model follows closely with the second-lowest bias. This negative bias likely reflects the tendency of NCAR's CMAQ field to underestimate ground-level concentrations, which our model may partially inherit through the proxy-consistency loss.

\Cref{tab:poverty_spatiotemporal_performance_summary} shows in-sample performance on the poverty mapping task.  As with PM$_{2.5}$, the proxy-only linear regression task performs poorly, suggesting that VIIRS night lights data alone is not a strong predictor of asset index.  However, naively stacking this data atop our Landsat observations provides a significant performance boost over a Landsat-only baseline, indicating the value of VIIRS data in the task.  By comparison, our frozen, pre-trained GeoCLIP \cite{vivanco2023geoclip} model performs poorly, perhaps due to the spatial extent of its Flickr pretraining dataset.  Unfreezing the location encoder does provide a performance boost, but still lags behind our proxy-stacked baseline.  Our PCL model outperforms both models by a substantial margin, across both $R^2$ and RMSE metrics.  We hypothesize that this gain is partly due to UAR proxy sampling, which may reduce overfitting to DHS training locations.

For air quality prediction, training with PCL adds ~3 s/epoch relative to the no-PCL fusion variant ($\approx$21 vs $\approx$18 s/epoch; \Cref{tab:app_pm25_mae_time}).  The observation encoder and stacked observation encoder take roughly 16 seconds per epoch, in comparison.
\begin{table}[t]
\centering
\small
\setlength{\tabcolsep}{3pt}
\begin{tabular}{lccc}
\hline
\textbf{Method} & $\boldsymbol{R^2}$ & \textbf{RMSE} \\
\hline
\multicolumn{3}{l}{\textit{Baselines}} \\
Proxy-only regression &$0.133$ & $1.573$ \\
Obs. encoder only &$-1.239\pm0.009$&$2.733\pm0.005$\\
Proxy-stacked &$\textit{0.366}\pm0.005$&$\textit{1.431}\pm0.006$\\
\hline
\multicolumn{3}{l}{\textit{Frozen location-encoder fusion}} \\
GeoCLIP&$-0.028\pm0.002$&$1.857\pm0.002$\\
Proxy pretraining (ours)&$-0.07972\pm0.001$&$1.9056\pm0.001$\\
\hline
\multicolumn{3}{l}{\textit{Trained location-encoder fusion}} \\
Without PCL &$0.241\pm0.004$&$1.583\pm0.004$\\
PCL ($\lambda=0.2$, $\rho = 1$) &$\textbf{0.462}\pm0.001$&$\textbf{1.343}\pm0.002$\\
\hline
\end{tabular}
\caption{\textbf{In poverty mapping}, PCL and heuristics targets in Africa with a spatio-temporal location encoder results in the best performance. Best results are in bold, second best are italicized.  Performance is averaged out over the course of 3 randomly seeded runs.  PCL results train on patch-level heuristics, predicting a tensor of min, max, mean, and centroid pixel values of a 128x128 VIIRS image patch centered on the sample's latitude and longitude.}
\label{tab:poverty_spatiotemporal_performance_summary}
\end{table}

\subsection{Out-of-sample Performance}

The out-of-sample results for air quality prediction, using a systematic offset checkerboard split with $\delta=8^{\circ}$ are given in \Cref{tab:pm2.5_uar_checkerboard_r2_rmse_mbe} and \Cref{tab:app_pm25_mae_time}. All methods experience performance degradation relative to the UAR spatial split, reflecting the challenge of spatial extrapolation. The proxy-only regressor remains weak, again highlighting the poor performance of the physical model and the scale mismatch between coarse reanalysis fields and point-level AQS targets. Proxy-stacking yields only a marginal improvement over the observation-only baseline.

Fusing frozen pretrained location encoders significantly enhances extrapolation. In this out-of-sample setting, GeoCLIP outperforms Climplicit, indicating that different pretraining targets provide varying levels of spatial generalizability. While training a location-time encoder from scratch without PCL fails to improve upon a proxy-stacked baseline, PCL-trained encoders consistently maintain the highest $R^{2}$ and lowest RMSE, demonstrating superior generalization to unseen regions. This corresponds to a $17.8\%$ gain in $R^{2}$ and a $4.4\%$ reduction in RMSE over the observation-only baseline. The near-zero MBE of the proxy-only regression should not be interpreted as low bias as its large SE is consistent with biases that flip sign across the swapped partitions and therefore cancel in the mean. Among the learned models, PCL yields the smallest MBE. 

\begin{figure}[t]
    \centering
    \includegraphics[width=1\columnwidth, trim=2mm 2mm 2mm 2mm,clip]{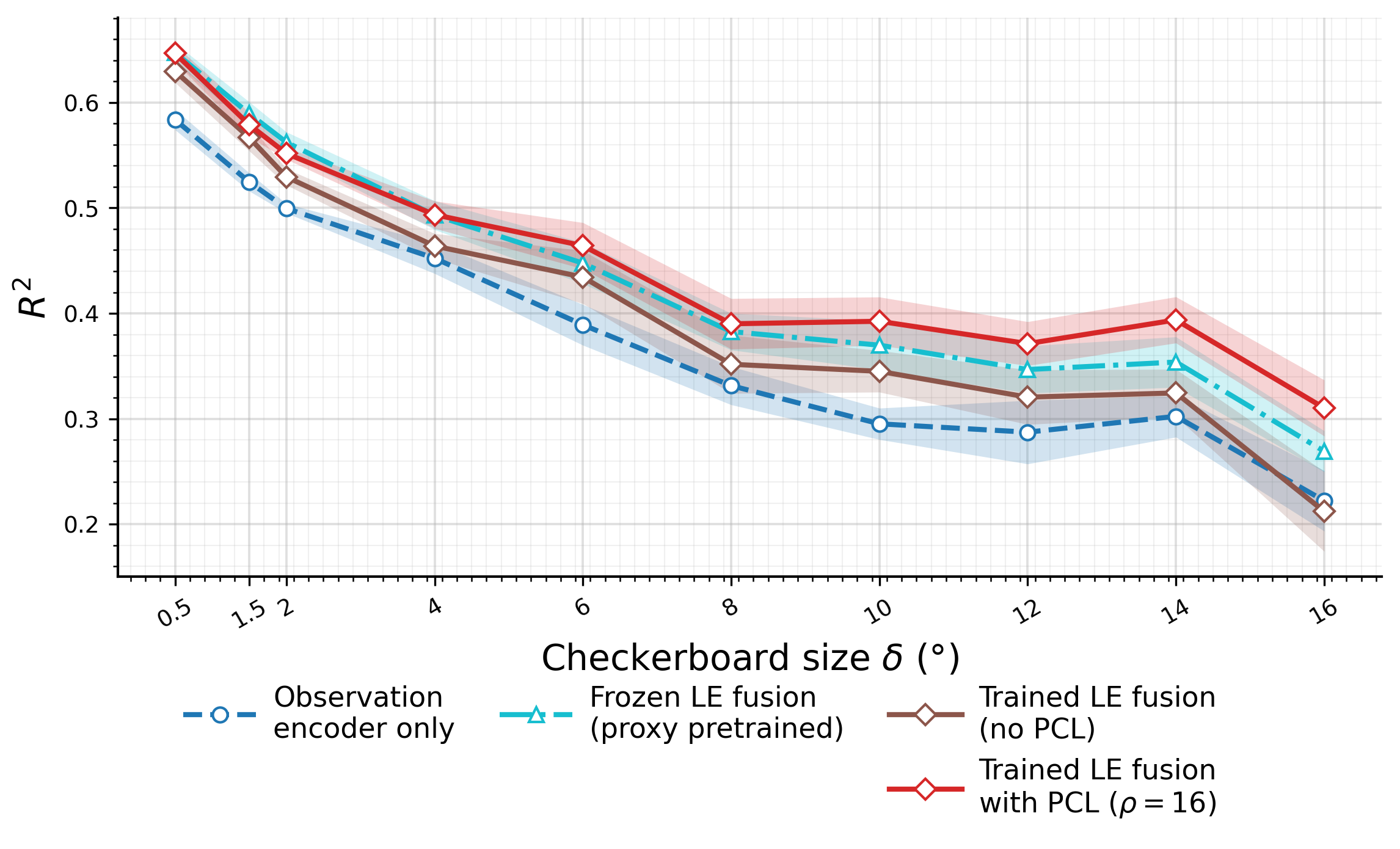}
    \caption{\textbf{Performance ($R^{2}$) on air quality task across checkerboard sizes $\delta$ ($^\circ$)}. As $\delta$ increases, performance generally degrades. Curves show the mean over 8 checkerboard partitions (4 spatial offsets $\times$ train/test swap), and shaded regions denote $\pm 1$ SE across partitions.}
    \label{fig:checkerboard-performance}
\end{figure}

\Cref{fig:checkerboard-performance} and \Cref{fig:checkerboard-performance-all} show the performance of our different methods as we vary the degree of spatial extrapolation as determined by the $\delta$ parameter. Performance degrades almost monotonically with larger $\delta$ as the gap between training and test regions widens. Across $\delta$, PCL-based fusion remains the top-performing method. Comparing joint PCL training to proxy-pretrained fusion, the two are nearly identical at small $\delta$ (and proxy-pretraining is slightly better at $\delta\in \{1.5^{\circ}, 2^{\circ} \}$ within uncertainty), while joint PCL becomes consistently better for larger $\delta$ (e.g., $\delta\ge 6^{\circ}$). This pattern suggests that while pretraining effectively captures fundamental spatial priors, joint training with PCL better regularizes the location-time representation and improves robustness under more severe extrapolation.

\begin{figure}[ht]
    \centering
    \includegraphics[width=1\columnwidth, trim=2mm 2mm 2mm 2mm,clip]{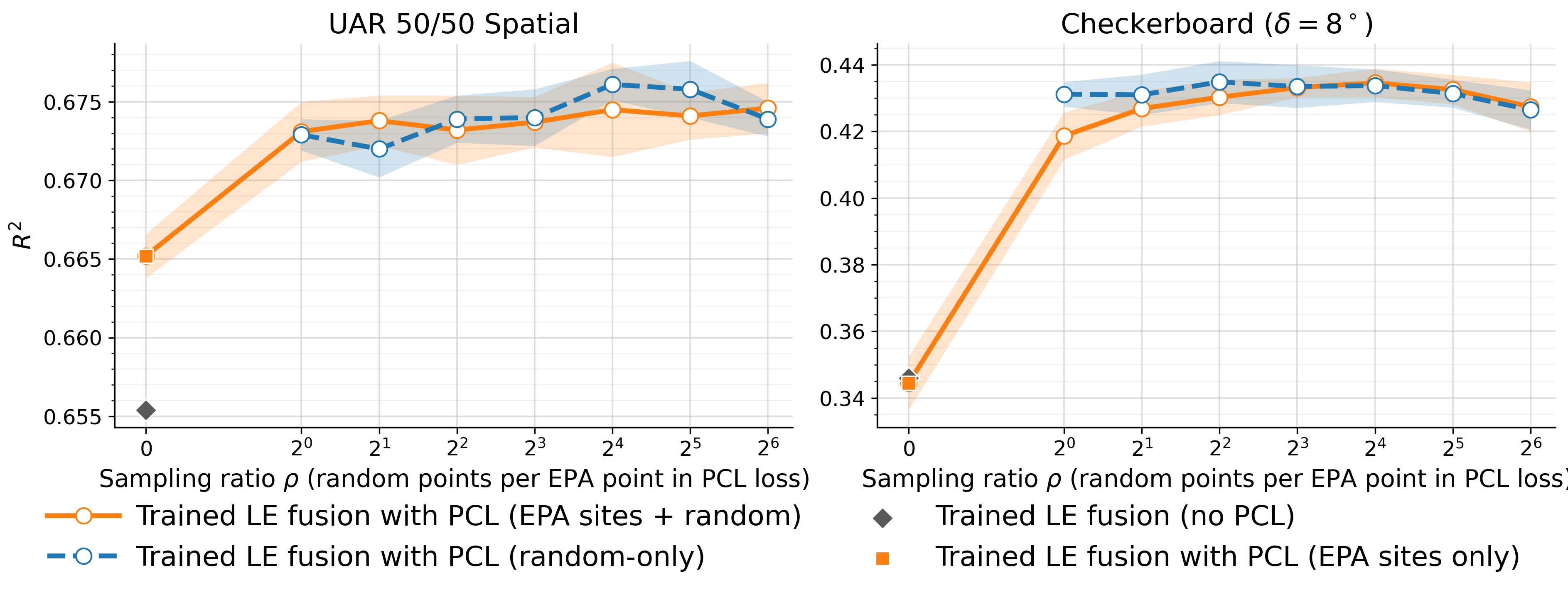}
    \caption{\textbf{PCL scaling with proxy sampling ratio $\rho$.} Performance of trained location-encoder fusion with proxy consistency loss (PCL) as we vary the sampling ratio $\rho$. Left: UAR 50/50 spatial split. Right: systematically offset checkerboard split ($\delta=8^\circ$). Solid curves report mean $R^2$ and shaded regions denote $\pm$1 standard error (SE) across runs/partitions. We compare two PCL sampling strategies: sampling proxy targets at EPA monitor coordinates in addition to $\rho$ uniformly sampled proxy points (PCL: EPA sites + random), versus using only uniformly sampled proxy points (PCL: random-only). Markers at $\rho=0$ indicate reference variants: trained LE fusion without PCL (No PCL) and PCL computed only at EPA monitor coordinates (PCL: EPA-only).}
    \label{fig:pcl_scaling}
\end{figure}

\subsection{Evaluating sampling decisions for PCL}
\label{sec:pcl_quant_analysis}
During training, we can sample batches of points to evaluate the PCL loss completely independently of where the primary target labels are available. \Cref{fig:pcl_scaling} shows how the amount and distribution of samples used to train the proxy consistency component of the model changes overall model performance. 

Colors in \Cref{fig:pcl_scaling} differentiate between whether the locations of the EPA sites are sampled to evaluate the PCL during training.
Under the spatial extrapolation setting, applying PCL only at EPA monitor locations (i.e., $\rho\!=\!0$ on the orange line) yields little to no benefit to not using the proxy variables at all, whereas adding proxy supervision at additional locations drives most of the gains. For in-sample prediction, using just the EPA station sites performs moderately well. In both settings, increasing the proxy sampling ratio $\rho$ improves performance rapidly at low-to-moderate values and then saturates around $\rho\!\approx\!8$--16, with slight degradation at very large $\rho$, potentially due to over-emphasizing the proxy objective when the proxy fields are imperfect approximations of the point-level target. Finally, it seems sampling proxy points uniformly over space-time is sufficient: random-only PCL (blue line) is competitive with (and sometimes slightly better than) sampling that also includes monitor locations. This suggests that broad proxy coverage rather than label-aligned proxy sampling is the key driver of improved geographic generalization.

\subsection{Qualitative Analysis and Additional Ablations} 
\subsubsection{Qualitative Analysis}
\label{sec:pcl_qual_analysis}
\begin{figure}[t]
\centering
    \begin{subfigure}[t]{0.49\linewidth}
        \centering\includegraphics[width=\linewidth]{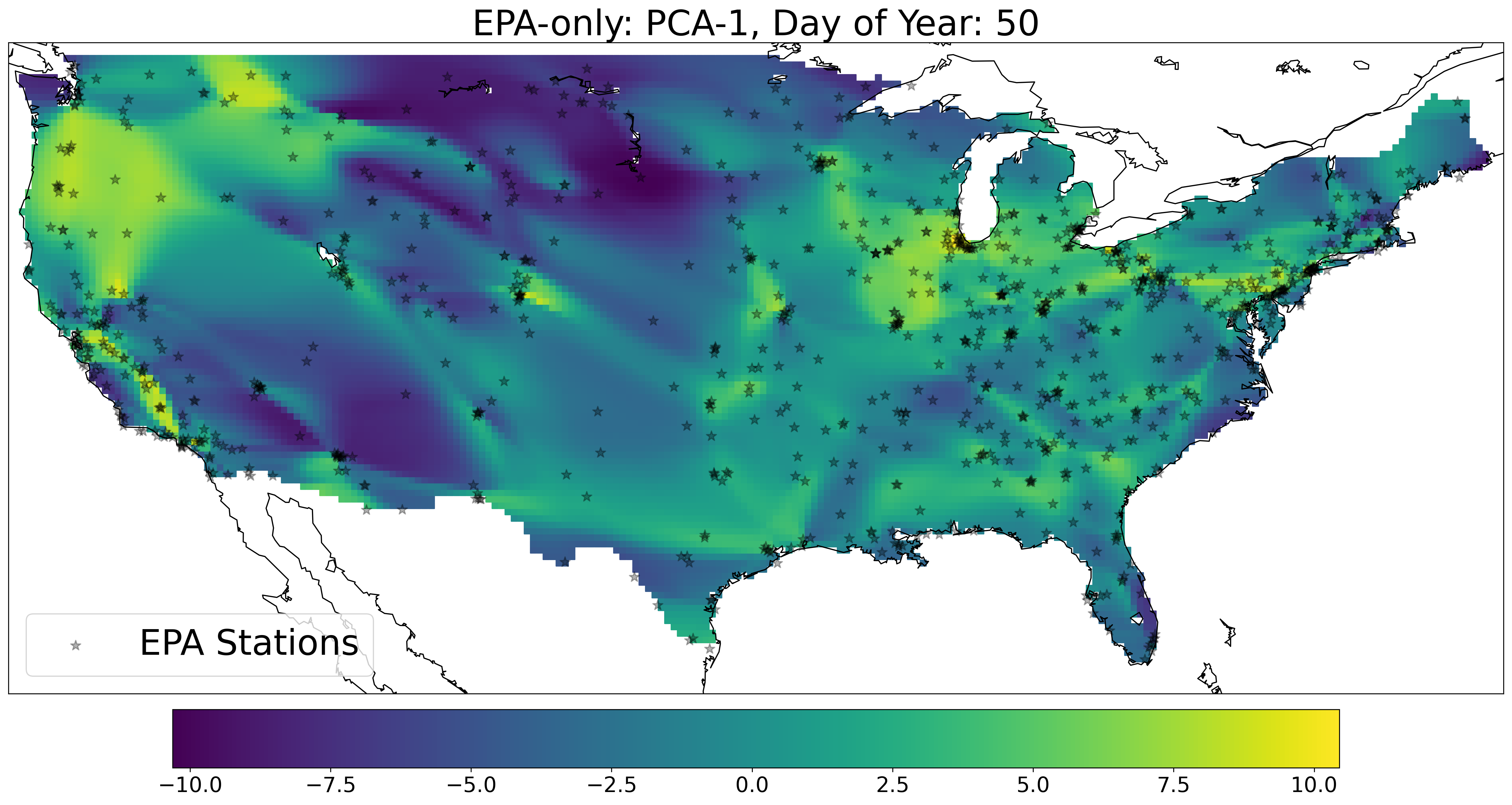}
    \end{subfigure}
    \begin{subfigure}[t]{0.49\linewidth}
        \centering\includegraphics[width=\linewidth]{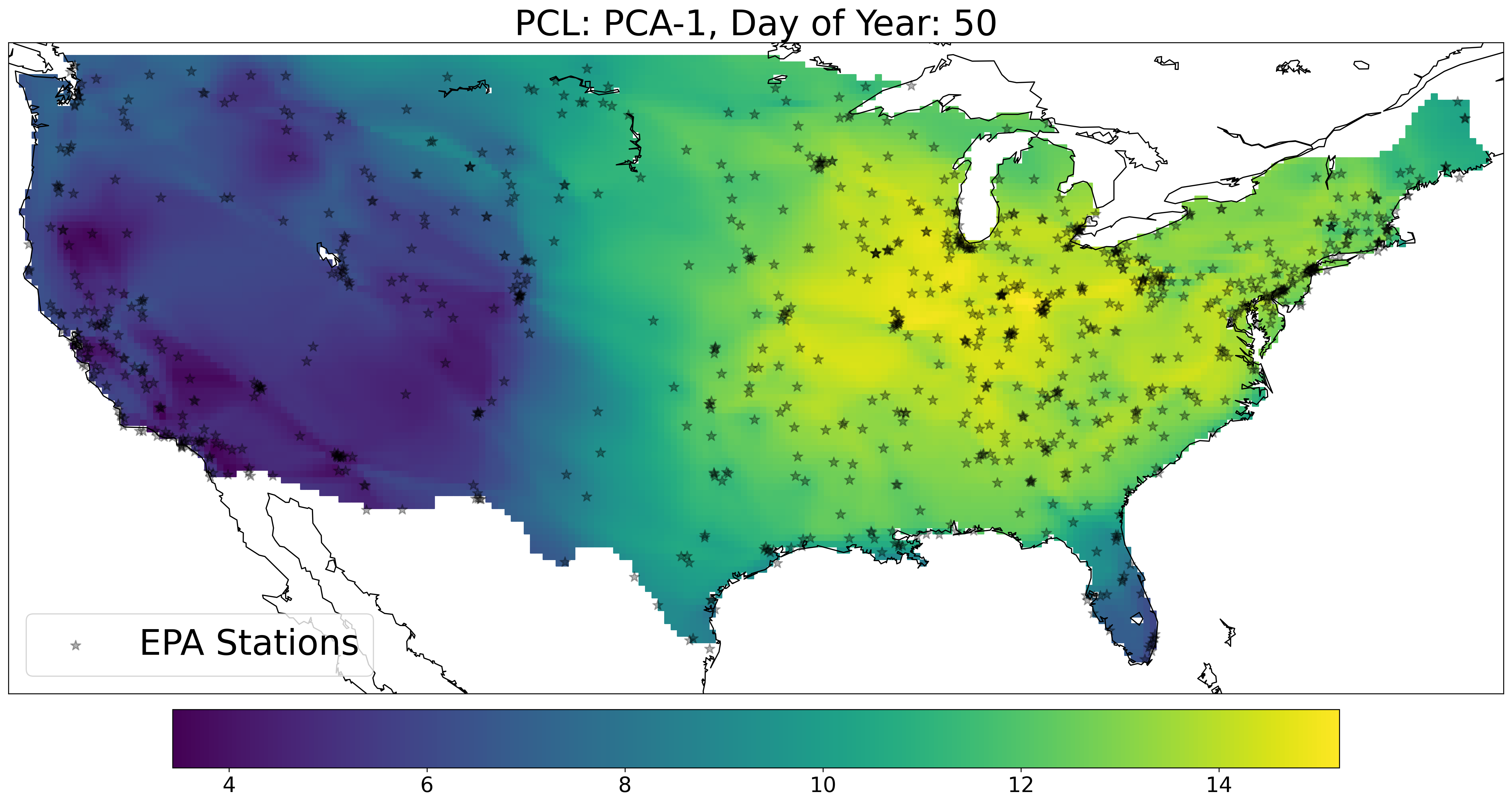}
    \end{subfigure}
    \begin{subfigure}[t]{0.49\linewidth}
        \centering
        \includegraphics[width=\linewidth]{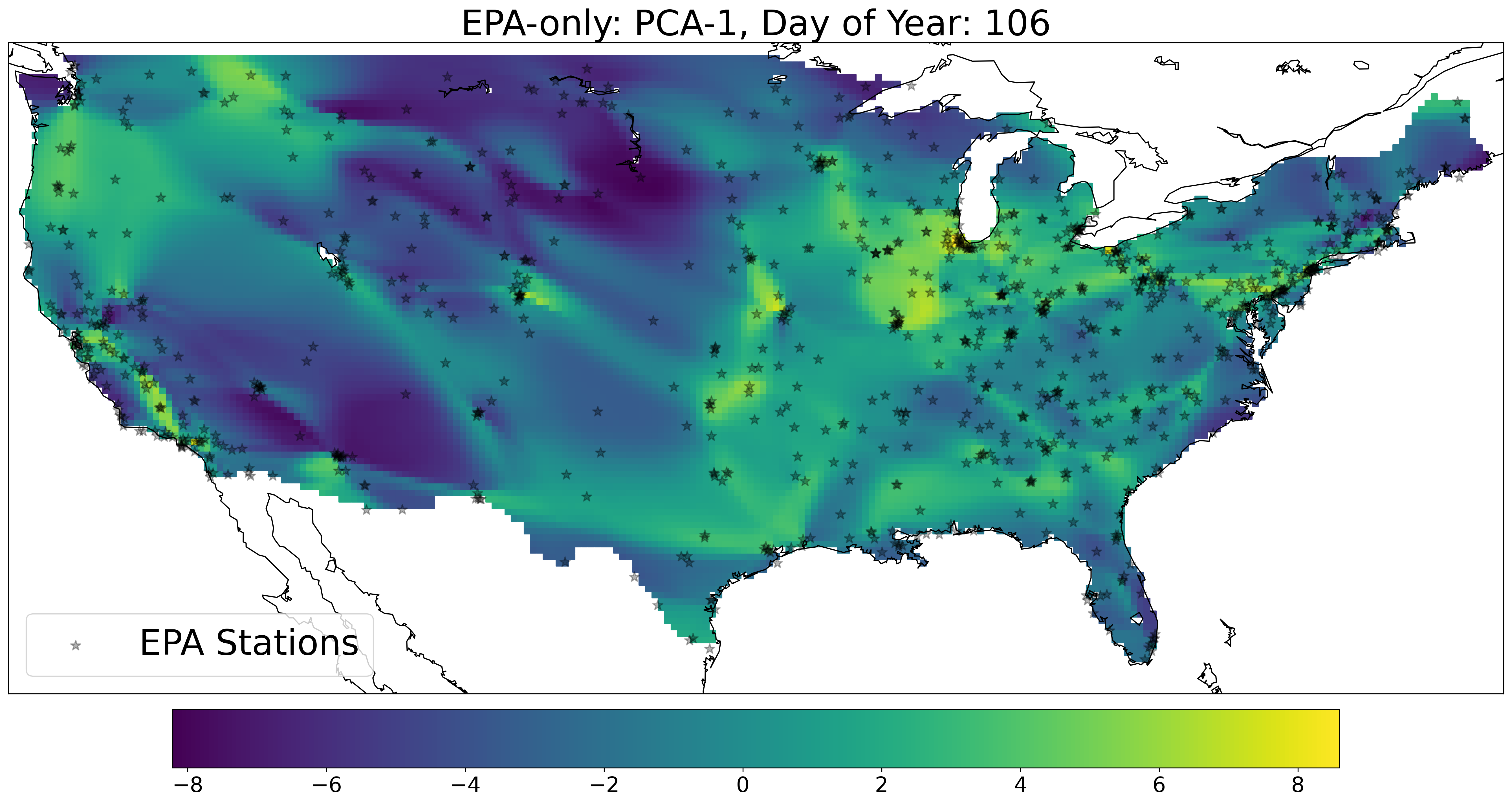}
    \end{subfigure}
        \begin{subfigure}[t]{0.49\linewidth}
        \centering\includegraphics[width=\linewidth]{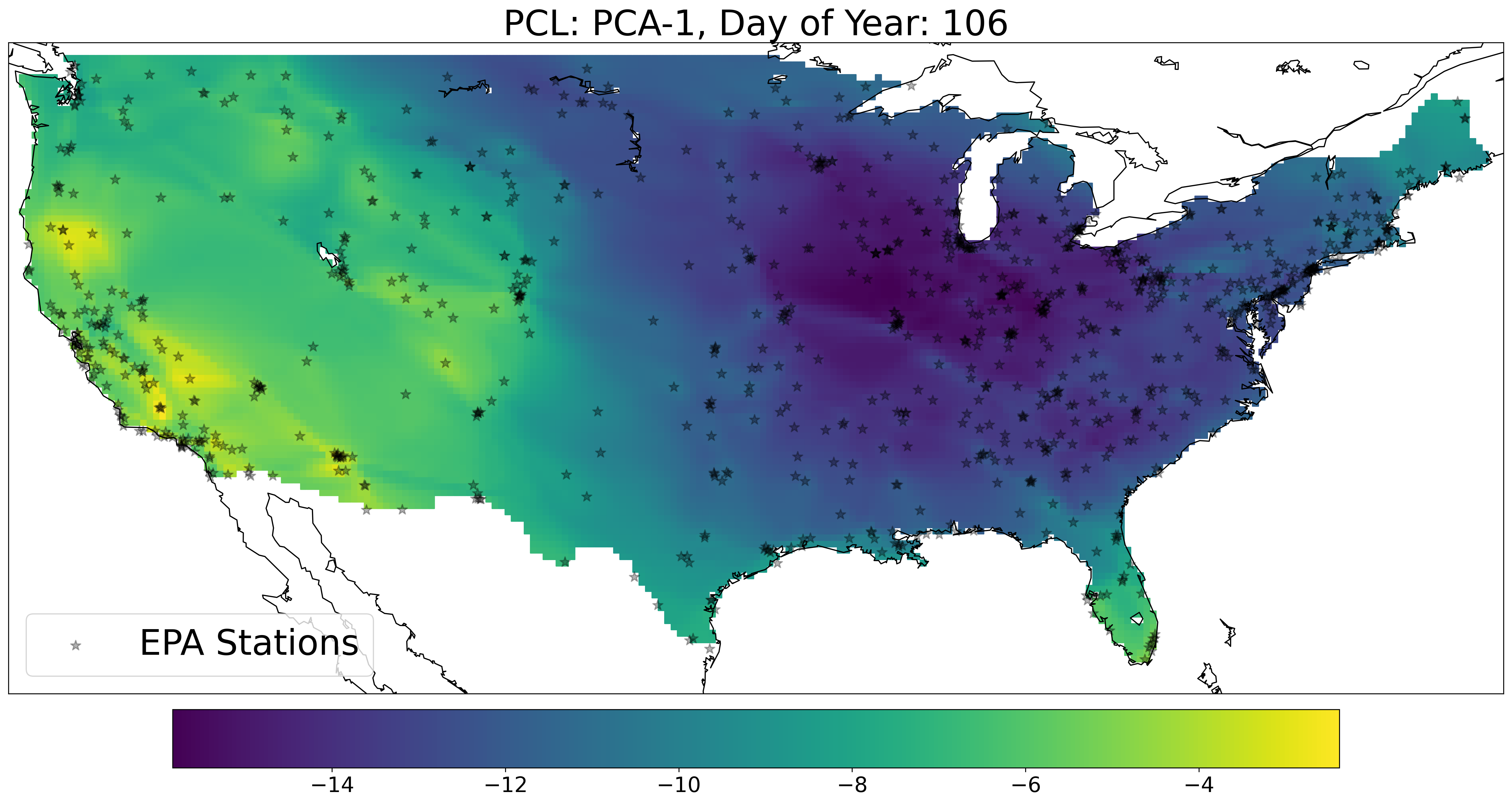}
    \end{subfigure}
    \begin{subfigure}[t]{0.49\linewidth}
        \centering
        \includegraphics[width=\linewidth]{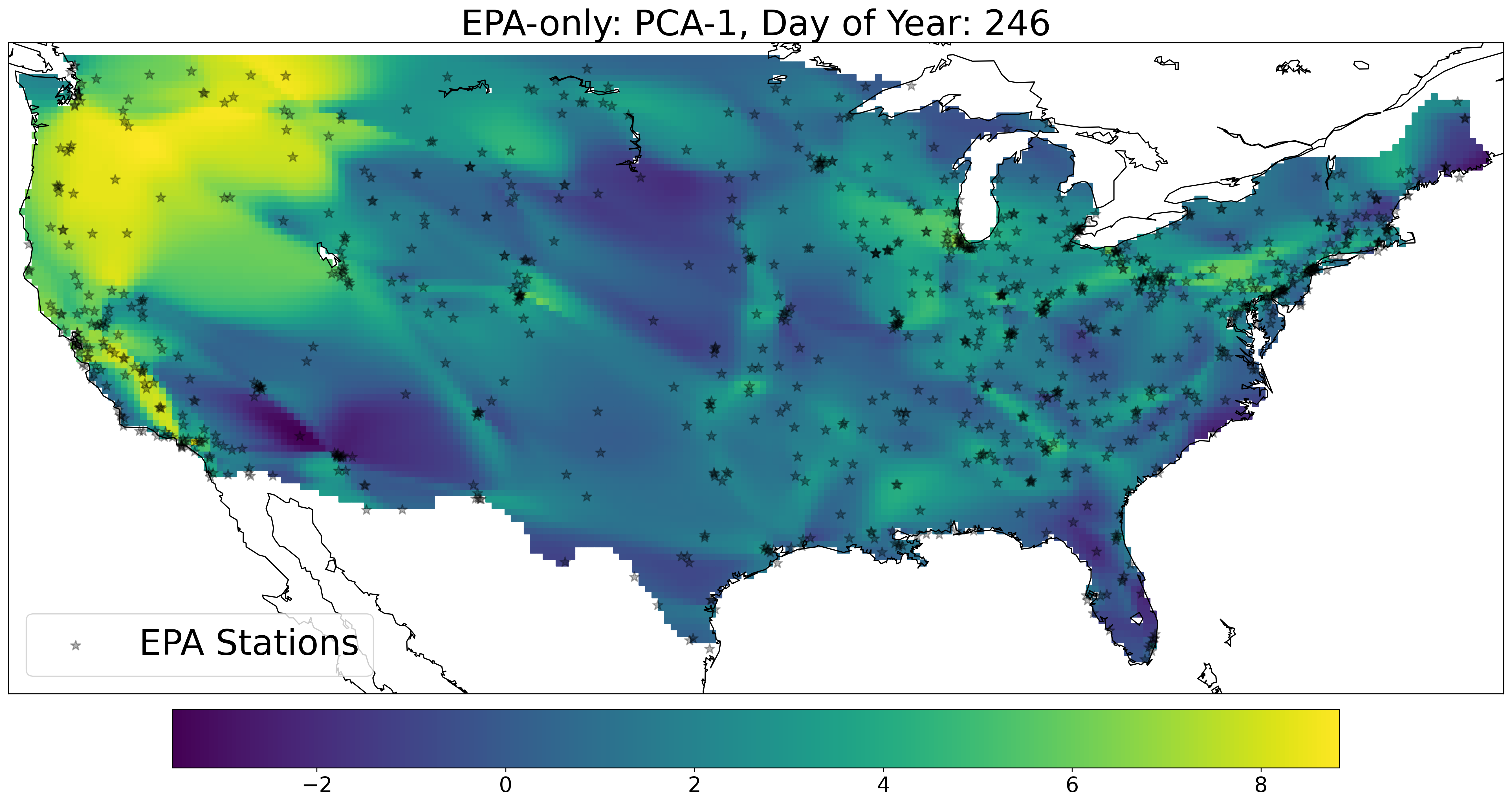}
    \end{subfigure}
    \begin{subfigure}[t]{0.49\linewidth}
        \centering\includegraphics[width=\linewidth]{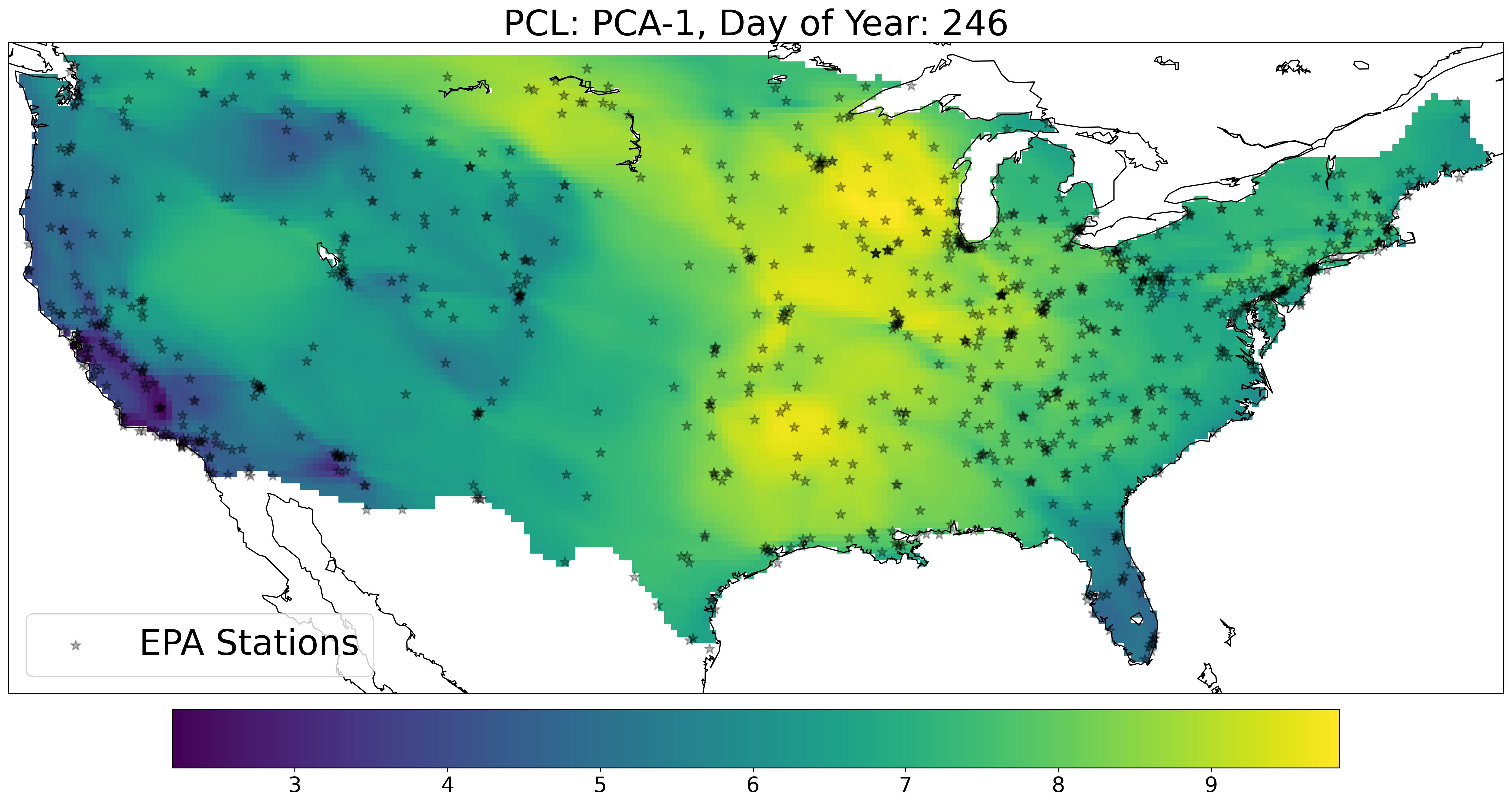}
    \end{subfigure}
    \caption{\textbf{In the air quality task} we show spatial impacts of EPA-only training (left) and PCL training (right) on different days-of-year. The EPA-only model overfits on EPA-station locations, whereas the PCL model is better regularized, with spatially smoother embedding features.}
    \label{fig:pm25_LE_spatial}
\end{figure}

\begin{figure}[h]
\centering
    \begin{subfigure}[t]{0.9\linewidth}
        \centering\includegraphics[width=\linewidth]{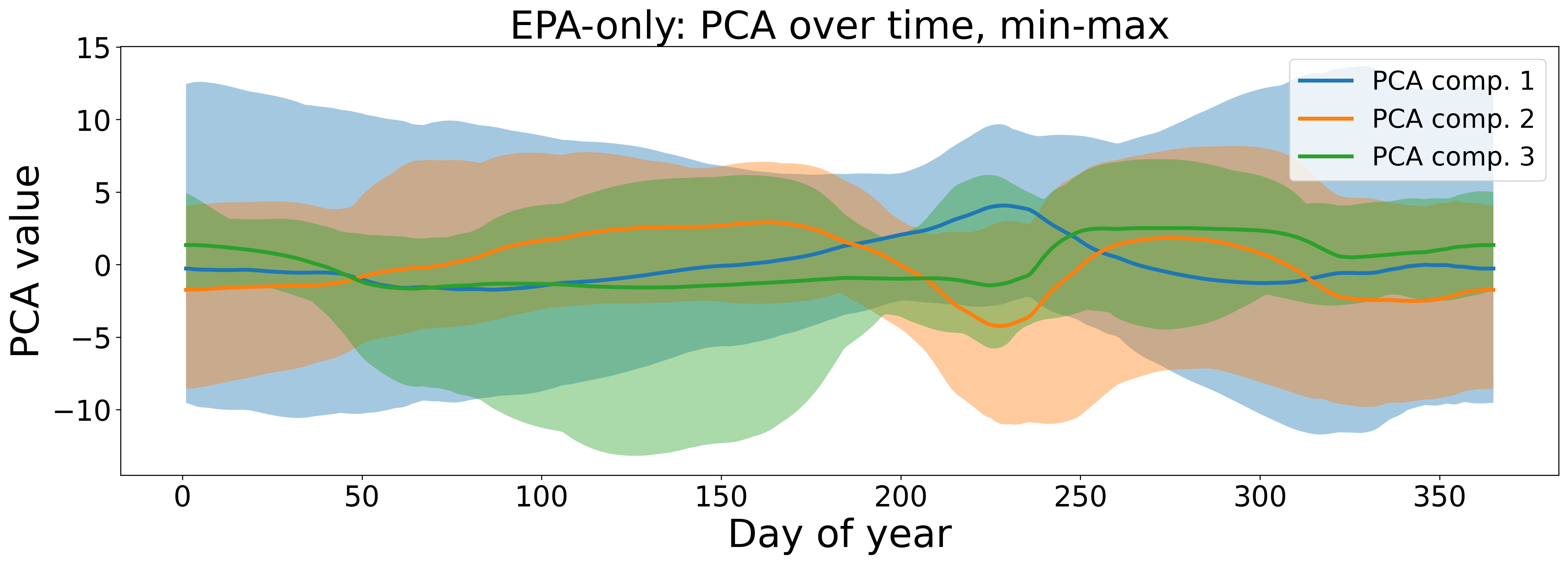}
    \end{subfigure}
    \begin{subfigure}[t]{0.9\linewidth}
        \centering
        \includegraphics[width=\linewidth]{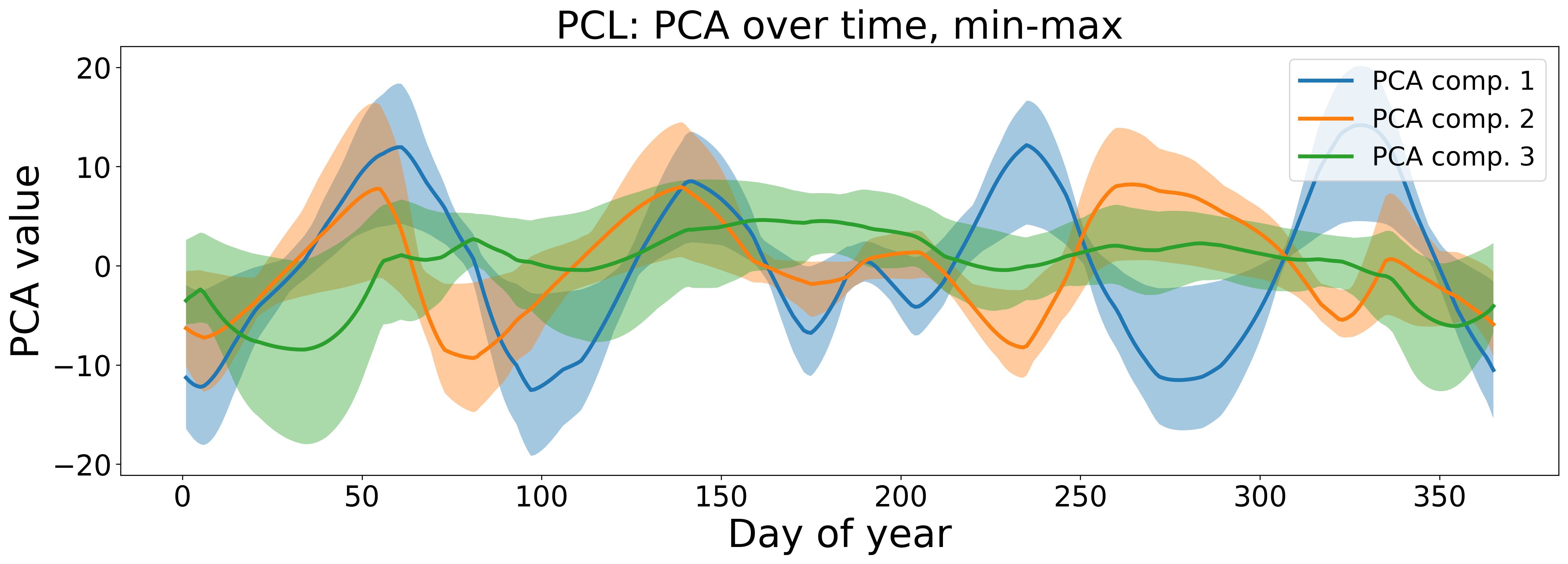}
    \end{subfigure}
    \caption{\textbf{In the air quality task} the EPA-only model fails to capture finer-grain temporal trends. The PCL-trained model instead shows temporal trends are dominant, as shown in \Cref{tab:pm2.5_uar_checkerboard_r2_rmse_mbe}, whereas the EPA-only model has larger spatial variance.}
    \label{fig:pm25_LE_temporal}
\end{figure}

To better understand how PCL changes the latent space of the location encoder in the fused prediction model, we visualize the location embeddings resulting from training the location encoder with the PCL and without it for the air quality prediction task. We extract embeddings evenly across a 0.25$^\circ$ grid of the continental US, and at daily time periods. We show the first PCA components of the extracted embeddings, plotted spatially, in \Cref{fig:pm25_LE_spatial}, which explains 13\% of the variance for the PCL training, and 4\% for the EPA-only. We also plot the temporal impacts in \Cref{fig:pm25_LE_temporal} by taking the mean across all embeddings in space at each time point; the minimum and maximum values of the respective PCA components are plotted to show the range.

Broadly, we see that embeddings resulting from training with PCL are spatially smoother than without (where we see artifacts stretching between EPA stations), which suggests that PCL regularizes training more appropriately. We also see that PCL has a stronger temporal structure, and that time dominates the variance, whereas EPA-only has little temporal structure, and is spatially more dominant. Interestingly, the PCL PCA values shift spatially over time in a bi-seasonal period. We also note that the EPA observations themselves when viewed naively, do not exhibit obvious temporal structure when averaged globally. But \Cref{tab:pcl_ablations} demonstrates the importance of the temporal regularization.

\label{subsec:pcl_ablations}

\begin{table*}[t]
\centering
\small
\setlength{\tabcolsep}{4pt}
\begin{tabular}{lccc c cc cc}
\toprule
\multirow{2}{*}{\textbf{PCL Variant}} &
\multirow{2}{*}{\textbf{Proxy}} &
\multirow{2}{*}{\textbf{Time}} &
\multirow{2}{*}{\textbf{Aux}} &
\multirow{2}{*}{\textbf{Sec/epoch}} &
\multicolumn{2}{c}{\textbf{UAR 50/50}} &
\multicolumn{2}{c}{\textbf{Checkerboard} $\boldsymbol{\delta=8^\circ}$} \\
\cmidrule(lr){6-7}\cmidrule(lr){8-9}
& & & & &
$\boldsymbol{R^2}$ & \textbf{RMSE} &
$\boldsymbol{R^2}$ & \textbf{RMSE} \\
\midrule
loc. = (lon, lat)                         & \checkmark &           &           & 20.2 & $0.652 \pm 0.002$ & $4.100 \pm 0.012$ & $0.387 \pm 0.020$ & $5.369 \pm 0.250$ \\
loc. = (lon, lat, DOY)                    & \checkmark & \checkmark &           & 21.3 & $0.671 \pm 0.001$ & $3.988 \pm 0.006$ & $0.390 \pm 0.024$ & $5.356 \pm 0.273$ \\
DOY + $\text{proxy}_{aux}$      &           & \checkmark & \checkmark & 21.4 & $0.663 \pm 0.002$ & $4.038 \pm 0.012$ & $0.355 \pm 0.025$ & $5.510 \pm 0.282$ \\
DOY + proxy + $\text{proxy}_{aux}$  & \checkmark & \checkmark & \checkmark & 21.9 & $\mathbf{0.675} \pm 0.003$ & $\mathbf{3.967} \pm 0.012$ & $\mathbf{0.398} \pm 0.024$ & $\mathbf{5.320} \pm 0.262$ \\
\bottomrule
\end{tabular}
\caption{\textbf{Ablations for trained LE fusion with proxy consistency loss (PCL, $\rho=16$).}
We vary (i) whether the primary proxy target (reanalysis total PM$_{2.5}$) is included in the proxy head, (ii) whether a day-of-year (DOY) temporal encoding is used to form a location-time encoder, and (iii) whether 7 additional auxiliary proxy targets from the NCAR reanalysis 
are included. We set the proxy and $\text{proxy}_{aux}$ loss weights to $\lambda_{\text{proxy}}=0.2$ and $\lambda_{\text{proxy}_{aux}}=0.4$, respectively. Results are reported as mean $\pm$ std. error across seeds/partitions (as in Table~\ref{tab:pm2.5_uar_checkerboard_r2_rmse_mbe}). 
}
\label{tab:pcl_ablations}
\end{table*}

\subsubsection{Ablations on air quality prediction} Time plays an important role in air quality dynamics. Adding a day-of-year temporal encoding (i.e., upgrading a pure location encoder to a location--time encoder) yields a clear gain under the UAR 50/50 spatial split (\Cref{tab:pcl_ablations}), but the improvement under checkerboard extrapolation is comparatively small. The second row in \Cref{tab:pcl_ablations} equates to the model used throughout the rest of the paper. 

Further ablations on proxy variable selection in \Cref{tab:pcl_ablations} show that augmenting PCL with additional auxiliary proxy targets from the same reanalysis (air density, relative humidity, surface temperature, planetary boundary layer height, solar radiation, precipitation, and 10-meter U/V-wind component) provides a stronger regularizer than using the main proxy (total PM$_{2.5}$) alone, particularly under larger spatial shifts. 
In contrast, using auxiliary reanalysis variables without the main PM$_{2.5}$ proxy improves UAR performance but degrades checkerboard extrapolation, indicating that proxy-consistency is most effective when it includes a task-aligned proxy rather than indirectly related variables.

\section{Discussion and Conclusion}

In cases where only sparse target labels are available, we propose a \textit{proxy consistency loss} as a strategy for incorporating supplemental data to construct robust, task-specific geographic priors. We showed that simply training location encoders on sparse observations may lead to unintentionally overfitting spatially and temporally, highlighted in the EPA-only results for the air quality task. At the same time, naïvely incorporating proxy data, as shown in the `stacked' baselines, only provides marginal gains in performance. Generally, location encoders aid in Earth observation-based prediction when they are regularized appropriately, as shown by the frozen encoder baselines and our PCL results. 

We note some limitations of this approach. First, it requires a proxy dataset that is related to the prediction task and as globally available as possible. While we showed two diverse settings where proxy data were readily available, in some cases, identifying good proxy datasets may be difficult. Second, encouraging latent-space consistency with proxy models may put the fused prediction models at-risk of learning representations that are more aligned to potential biases in the proxy data than would occur without the PCL objective. For example, in the air quality prediction task, if the reanalysis underestimates PM$_{2.5}$ concentrations during unmeasured instances of fire, it may bias the overall model to underestimate these regions or periods as well. So balancing these and selecting an appropriate $\lambda$ may be difficult, and will be the subject of future work.

Overall, we found that by incorporating proxy and auxiliary data as a secondary prediction problem, our proposed PCL maintains geospatial and temporal consistency and results in the best performance for in-sample and out-of-sample prediction of all the methods we compared. The encoders that are trained on the additional proxy data demonstrate more coherent spatiotemporal structure when visualized, affirming our intuition that the latent space of location encoders can be effectively shaped for location-aware supervised learning with Earth observation inputs.

\section*{Acknowledgments}
This research was supported by the CU Boulder Research \& Innovation Office and the NSF award \# DBI 2153040 (ESIIL).

\bibliographystyle{ieeenat_fullname}
\bibliography{main}

\appendix
\clearpage
\setcounter{page}{1}
\maketitlesupplementary

\appendix

\renewcommand{\thetable}{S\arabic{table}}
\renewcommand{\thefigure}{S\arabic{figure}}

\crefname{table}{Table}{Tables}
\Crefname{table}{Table}{Tables}
\crefname{figure}{Figure}{Figures}
\Crefname{figure}{Figure}{Figures}

\setcounter{table}{0}
\setcounter{figure}{0}

\section{Air quality prediction details}

\subsection{Dataset details}
\label{sec:appendix_aqp_dataset}
Earth observation features used in the air quality prediction task include time series of: Moderate Resolution Imaging Spectroradiometer (MODIS) Multi-Angle Implementation of Atmospheric Correction (MAIAC) aerosol optical depth (AOD) at 0.47 and 0.55 $\mu m$, Daymet meteorology (day length, precipitation, shortwave radiation, min/max temperature, vapor pressure), gridMET near-surface winds (direction and speed), MODIS Normalized Difference Vegetation Index (NDVI), Global Multi-resolution Terrain Elevation Data 2010 (GMTED2010) elevation, and wildfire smoke density from the National Oceanic and Atmospheric Administration (NOAA) Hazard Mapping System (HMS) Smoke product \cite{wang2025high}.

The reanalysis used for the air quality prediction proxy dataset is generated by assimilating MODIS AOD and Measurement of Pollution in the Troposphere (MOPITT) carbon monoxide (CO) retrievals into the Community Multiscale Air Quality (CMAQ) chemical transport model (CTM), driven offline by Weather Research and Forecasting (WRF) meteorology, using a three-dimensional variational (3D-Var) approach \cite{kumar2025long}. 

\subsection{Training details}

\begin{figure*}[t]
    \centering
    \begin{subfigure}[t]{0.48\textwidth}
        \centering
        \includegraphics[width=\linewidth]{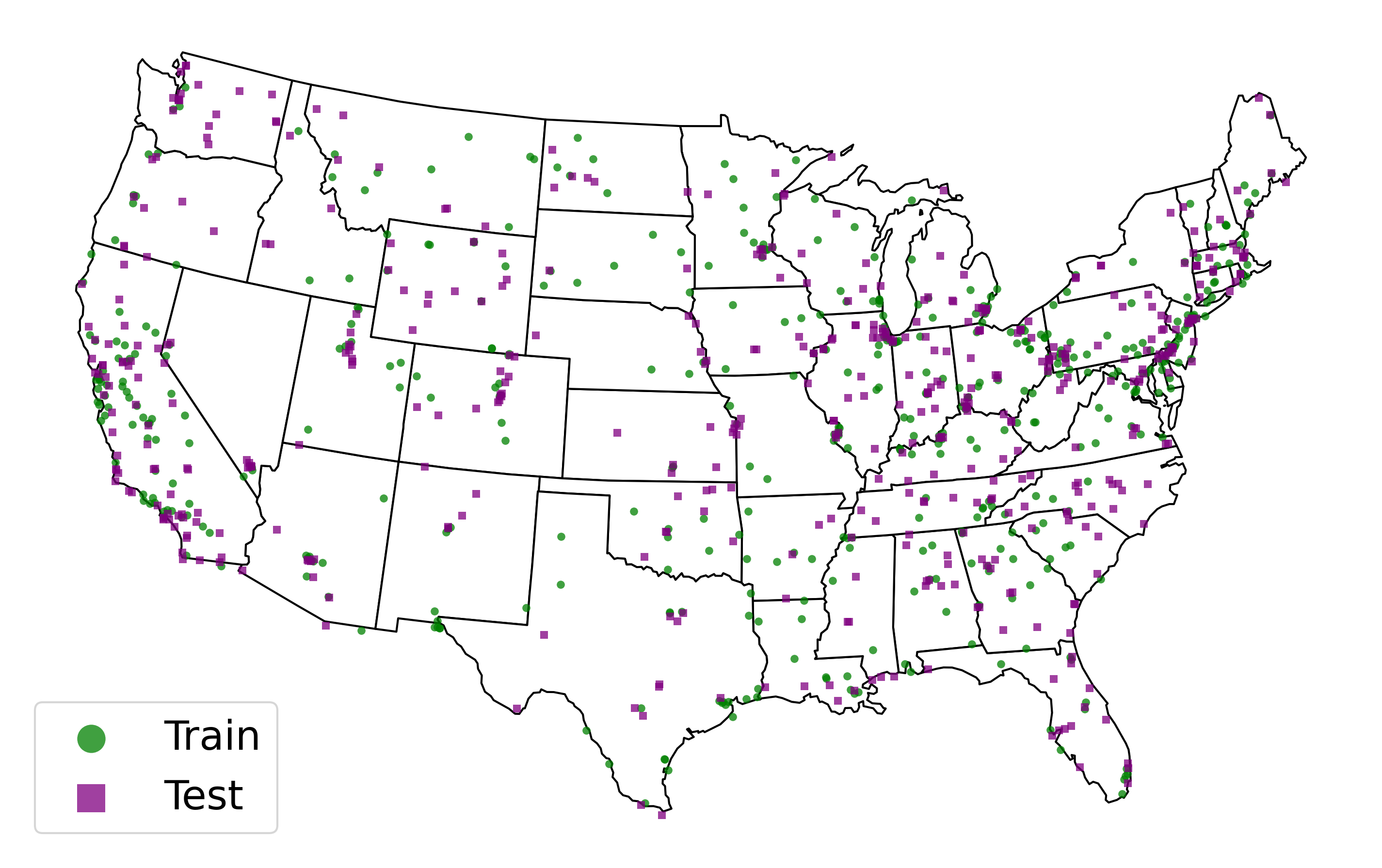}
        \caption{UAR 50/50 split for PM$_{2.5}$ prediction task.}
        \label{fig:epa-uar-split}
    \end{subfigure}\hfill
    \begin{subfigure}[t]{0.48\textwidth}
        \centering
        \includegraphics[width=\linewidth]{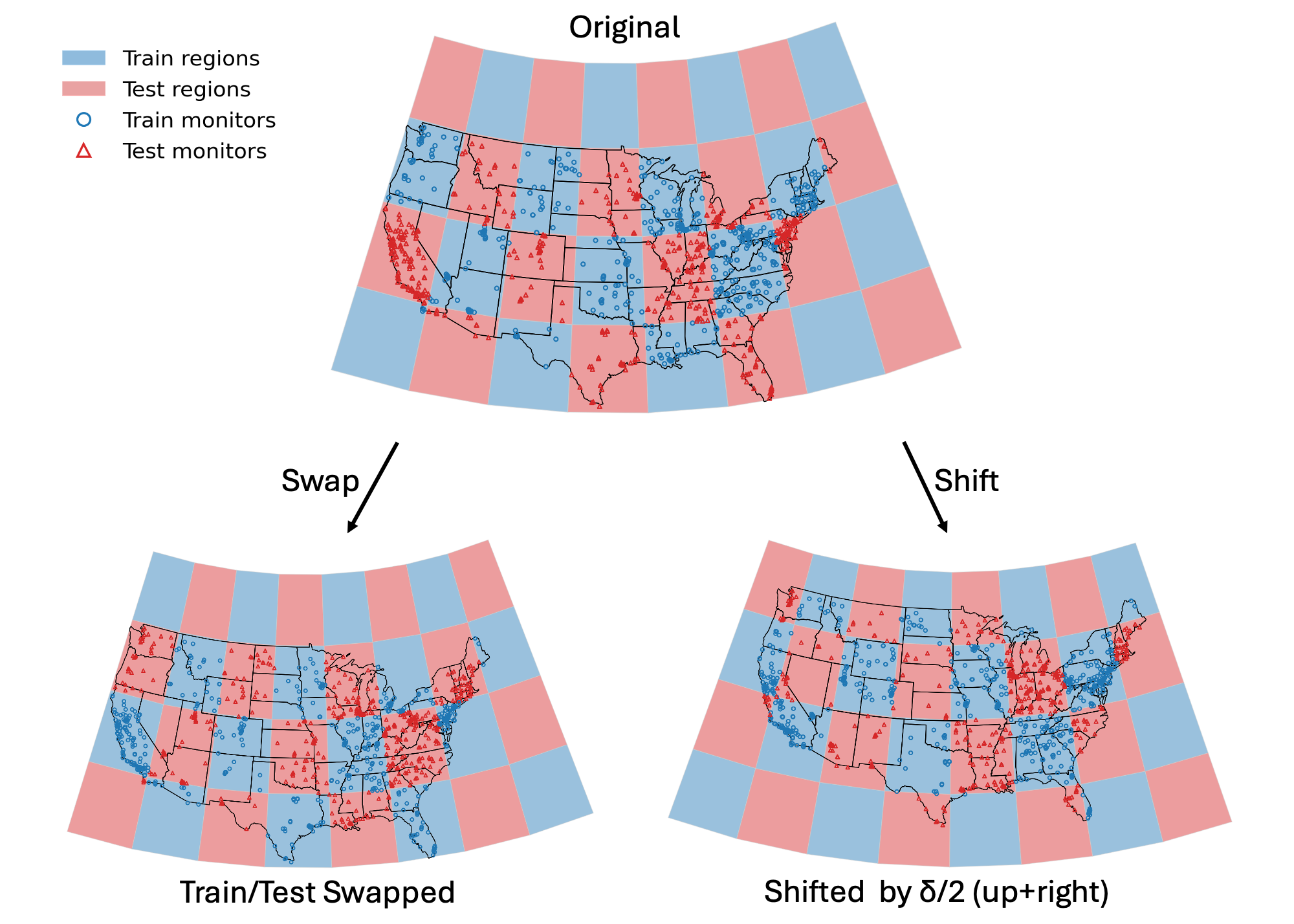}
        \caption{Systematic checkerboard split ($\delta=8^{\circ}$) and augmentations used to measure variation in results.}
        \label{fig:epa-checkerboard-split}
    \end{subfigure}
    \caption{\textbf{Geographic in-sample and out-of-sample evaluation protocols for the PM$_{2.5}$ prediction task.} Plotted points represent EPA stations where in-situ PM$_{2.5}$ measurements are taken. (a) Uniform-at-random (UAR) 50/50 location split, where test sites may be geographically close to training sites. (b) Systematic checkerboard split for geographic extrapolation. We evaluate 8 partitions per $\delta$ (4 checkerboard offsets $\times$ train/test swap).}
    \label{fig:epa-splits}
\end{figure*}

\Cref{fig:epa-checkerboard-split} shows the process for generating the spatial train/validation and test splits.
To capture variation to the placement of grid boundaries, we repeat this procedure under four checkerboard offsets (original, shifted right by $\delta/2$, shifted up by $\delta/2$, and shifted both up and right by $\delta/2$). In addition, for each offset, we also swap the train/test squares, yielding 8 partitions in total (4 offset $\times$ train/test swap).

Unless otherwise noted, each optimization step samples a minibatch of $B=256$ labels and an additional minibatch of $\rho$ proxy samples drawn uniformly over space-time (default $\rho=16$). All models are trained with AdamW \cite{loshchilov2017decoupled} for 100 epochs (learning rate $3{\times}10^{-4}$; gradient clipping at 1.0) with ReduceLROnPlateau scheduling and early stopping.

For proxy pretraining, we first pretrain only the location-time encoder (together with a proxy-prediction head) for 50 epochs on the proxy prediction task, using the same optimizer and hyperparameter settings as in the main experiments. The Earth observation encoder is not used during this pretraining stage and is initialized randomly for the downstream task. After pretraining, we discard the proxy-prediction head, freeze the pretrained location-time encoder, and train the Earth observation encoder together with the downstream prediction head for 100 epochs. Thus, all methods that use a frozen location encoder (including GeoCLIP, Climplicit, and proxy pretraining) receive the same 100-epoch training. We include this baseline to isolate the value of task-aligned proxy supervision without joint end-to-end fusion.

\section{Poverty mapping task details}
\subsection{Training details}
Unless otherwise specified, we train each model for 50 epochs with a learning rate of 0.0001 and a batch size of 128, using an AdamW optimizer \cite{loshchilov2017decoupled}.  Aside from the loading of pre-trained weights outlined above, all modules are trained simultaneously.  For models with a proxy task, we use multiple optimizers to propagate loss through the appropriate parameters.




\section{Detailed air quality prediction results}
\Cref{tab:app_pm25_mae_time} supplements the results of \Cref{tab:pm2.5_uar_checkerboard_r2_rmse_mbe} in the main text.

\begin{table*}[t]
\centering
\small
\setlength{\tabcolsep}{4pt}
\begin{tabular}{lcc cc c}
\toprule
\textbf{Method} &
\multicolumn{2}{c}{\textbf{UAR 50/50}} &
\multicolumn{2}{c}{\textbf{Checkerboard} ($\boldsymbol{\delta}=8^{\circ}$)} &
\textbf{Time/epoch} (s) \\
\cmidrule(lr){2-3}\cmidrule(lr){4-5}
& \textbf{MAE} (\si[per-mode=symbol]{\micro\gram\per\cubic\meter}) &
&
\textbf{MAE} (\si[per-mode=symbol]{\micro\gram\per\cubic\meter}) &
&
\\
\midrule
\multicolumn{6}{l}{\textit{Baselines}} \\
Proxy-only regression & 3.174 & & $3.200 \pm 0.041$ & & -- \\
Obs. encoder only & $2.428 \pm 0.003$ & & $3.087 \pm 0.074$ & & 15.85 \\
Proxy-stacked obs. encoder & $2.388 \pm 0.005$ & & $2.974 \pm 0.074$ & & 16.25 \\
\midrule
\multicolumn{6}{l}{\textit{Frozen location-encoder fusion}} \\
GeoCLIP & $2.360 \pm 0.003$ & & \textit{2.923} $\pm$ 0.068 & & 17.55 \\
Climplicit & $2.286 \pm 0.003$ & & $3.062 \pm 0.072$ & & 17.52 \\
Proxy pretraining (ours) & \textit{2.190} $\pm$ 0.006 & & $2.997 \pm 0.075$ & & -- \\
\midrule
\multicolumn{6}{l}{\textit{Trained location-encoder fusion}} \\
without PCL & $2.239 \pm 0.005$ & & $2.991 \pm 0.089$ & & 18.40 \\
PCL ($\lambda=0.2$, $\rho = 16$) & $\mathbf{2.186} \pm 0.003$ & & $\mathbf{2.857} \pm 0.078$ & & 21.29 \\
\bottomrule
\end{tabular}
\caption{\textbf{Additional metrics and efficiency (MAE and time/epoch) for air quality task.}
MAE is reported as mean $\pm$ SE over 5 random seeds for UAR and over 8 partitions (4 offsets $\times$ train/test swap) for checkerboard. Time/epoch is measured on the same hardware as the main experiments and is not reported for proxy-pretrained fusion due to two-stage training. Corresponding $R^2$, RMSE, and MBE are reported in Table~\ref{tab:pm2.5_uar_checkerboard_r2_rmse_mbe}.}
\label{tab:app_pm25_mae_time}
\end{table*}

\begin{figure*}[t]
    \centering
    \includegraphics[width=1\linewidth, trim=8 6 8 6,clip]{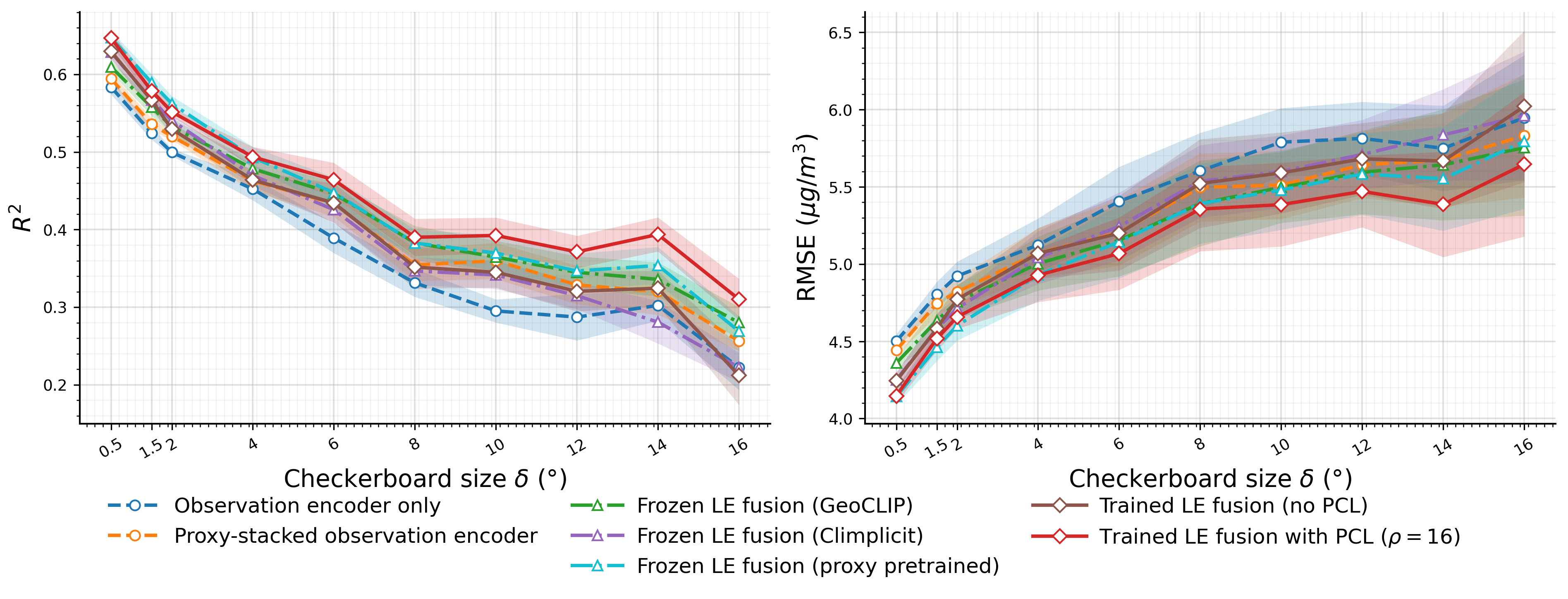}
    \caption{\textbf{Air quality prediction performance across checkerboard sizes $\delta$ ($^\circ$)}. \textit{Left:} $R^{2}$ (higher is better). \textit{Right:} RMSE (lower is better). As $\delta$ increases (harder spatial extrapolation), performance generally degrades (decreasing $R^{2}$ and increasing RMSE) across all methods. Curves show the mean over 8 checkerboard partitions (4 spatial offsets $\times$ train/test swap), and shaded regions denote $\pm 1$ SE across partitions.}
    \label{fig:checkerboard-performance-all}
\end{figure*}

\end{document}